\documentclass[letterpaper, 10 pt, conference]{ieeeconf}  
\usepackage{amsmath}
\usepackage{amsfonts}
\usepackage{amssymb}
\usepackage{graphicx}
\usepackage{siunitx}
\usepackage{subfig}
\usepackage{booktabs} 
\usepackage{multirow} 
\usepackage{tabularx} 
\usepackage[export]{adjustbox} 
\usepackage{url}
\usepackage{bm}  

\IEEEoverridecommandlockouts                              
\title{\LARGE \bf
Robust Co-design Optimisation for Agile Fixed-Wing UAVs}

\author{Adrian A. Buda$^{1}$, Xavier Chen$^{1}$, Nicolò Botteghi$^{2}$, and Urban Fasel$^{1}$
\thanks{$^{1}$Department of Aeronautics, 
   Imperial College London, United Kingdom
}
\thanks{$^{2}$Department of Mathematics, Politecnico di Milano, Italy 
}
\thanks{Source code: {\scriptsize \protect\url{https://github.com/adrianbuda30/robust_UAV}}}
\thanks{\copyright~2026 IEEE. Personal use of this material is permitted.
Permission from IEEE must be obtained for all other uses, in any current or
future media, including reprinting/republishing this material for advertising
or promotional purposes, creating new collective works, for resale or
redistribution to servers or lists, or reuse of any copyrighted component of
this work in other works. This paper has been published in the \emph{2026 International Conference on
Unmanned Aircraft Systems (ICUAS)}.}
}

\begin{document}

\maketitle

\thispagestyle{empty}
\pagestyle{empty}

\begin{abstract}

Co-design optimisation of autonomous systems has emerged as a powerful alternative to sequential approaches by jointly optimising physical design and control strategies. However, existing frameworks often neglect the robustness required for autonomous systems navigating unstructured, real-world environments. For agile Unmanned Aerial Vehicles (UAVs) operating at the edge of the flight envelope, this lack of robustness yields designs that are sensitive to perturbations and model mismatch. To address this, we propose a robust co-design framework for agile fixed-wing UAVs that integrates parametric uncertainty and wind disturbances directly into the concurrent optimisation process. Our bi-level approach optimises physical design in a high-level loop while discovering nominal solutions via a constrained trajectory planner and evaluating performance across a stochastic Monte Carlo ensemble using feedback LQR control. Validated across three agile flight missions, our strategy consistently outperforms deterministic baselines. The results demonstrate that our robust co-design strategy inherently tailors aerodynamic features, such as wing placement and aspect ratio, to achieve an optimal trade-off between mission performance and disturbance rejection.

\end{abstract}

\section{INTRODUCTION}

Fixed-wing Unmanned Aerial Vehicles (UAVs) have a broad range of applications, from long-endurance surveillance missions to large-scale geospatial mapping, precision agriculture, and critical infrastructure monitoring tasks \cite{Mitridis}. Recent literature has increasingly focused on developing more accurate aerodynamic models, planning, and control strategies that enable agile flight by handling nonlinearities and instabilities more effectively when flying beyond the nominal flight envelope \cite{Khan}\cite{Tal}\cite{Coletti}. These advancements allow deploying fixed-wing UAVs to complex, highly constrained environments, such as search and rescue missions and high-speed urban navigation tasks \cite{Basescu}, where agile multi-rotor platforms have already demonstrated remarkable performance \cite{Kaufmann}. However, pushing conventional fixed-wing UAVs to fly aggressive, highly dynamic manoeuvres is challenging, since they are designed to be inherently stable and naturally counteract sudden changes in orientation and disturbances. Avian-inspired morphing wings offer a promising solution to adapt the morphology to different flight regimes, either actively or passively, by altering the stability characteristics and enhancing agility and manoeuvrability \cite{Harvey}\cite{Ajanic}. Although active morphing mechanisms have been successfully integrated and flight-tested on different winged UAV platforms \cite{fasel2020composite}\cite{Wuest}, the coupling between flight dynamics, unsteady aerodynamics, and structural deformation makes these systems difficult to control in a robust manner. 

Conventionally, fixed-wing UAVs follow a sequential design process: the airframe is designed and configured for a set of pre-defined mission constraints, then a suitable control strategy is developed to achieve a desired task \cite{Beard}. This sequential design and control paradigm can yield suboptimal results, particularly in agile flight regimes. An alternative approach is to treat the hardware and controller as a unified system that is jointly optimised to maximise global mission performance \cite{Digumarti}. This co-design strategy can concurrently optimise the physical design and control strategy, viewed as a coupled system to achieve optimal design-control pairs for specific missions. The paradigm was initially proposed in Evolutionary Robotics to exploit the strong interconnection between a robot's "body" or hardware and its "brain" or controller \cite{Sims}. Since then, co-design optimisation has been successfully applied across a wide range of robotic platforms, from legged robots \cite{Digumarti}\cite{Ha}\cite{Schaff}\cite{Palacios}\cite{Fadini} and manipulators \cite{Spielberg}\cite{Ciocarlie} to flying robots such as multi-rotors \cite{Du}\cite{Carlone} and fixed-wing UAVs \cite{Bergonti}\cite{Zhao}.

\begin{figure*}[t] 
    \centering
    \includegraphics[width=1.0\textwidth]{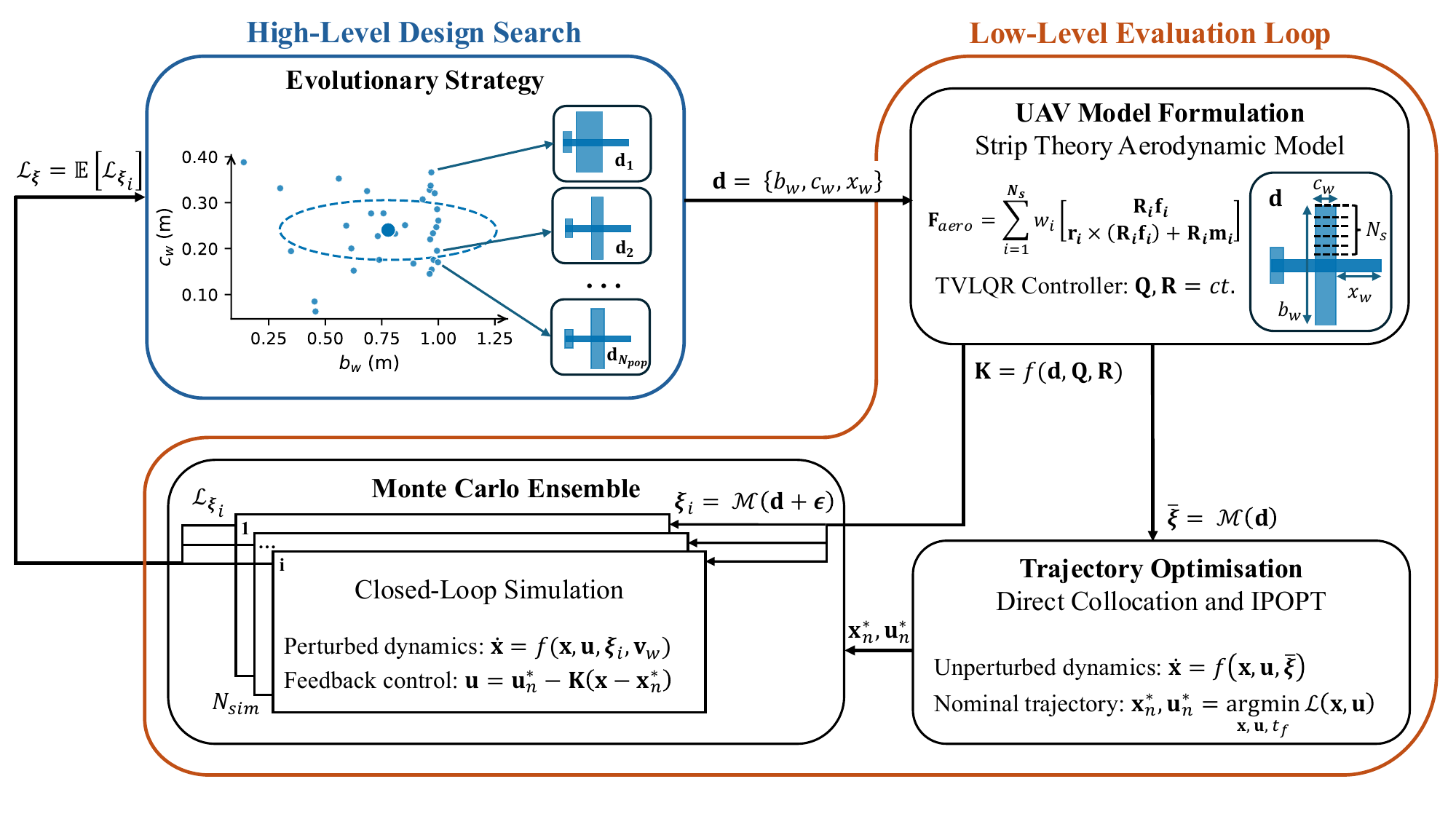}
    \caption{Overview of the bi-level robust co-design framework. Candidate designs $\mathbf{d} \in \mathcal{D}$ are sampled by CMA-ES (the high-level design search loop) and evaluated through a low-level loop consisting of: (i) the UAV model and Time-Varying LQR controller formulation; (ii) the trajectory optimisation via direct collocation using the nominal system dynamics; and (iii) the Monte Carlo ensemble of stochastic closed-loop simulations. The robust global optimisation task seeks the jointly optimal set of design parameters $\mathbf{d}^{*}$ and controlled trajectory $(\mathbf{x}^{*}, \mathbf{u}^{*})$ that minimise the aggregate cost $\mathcal{L}_{\boldsymbol{\xi}}$ across the $N_{sim}$ simulations.}
    \label{overview}
    \vspace{-12pt}
\end{figure*}

Most co-design frameworks focus on nominal performance and do not explicitly account for robustness or the inherent challenges of sim-to-real transfer \cite{Scaramuzza}. While directly considering this in the co-design optimisation remains rare for flight systems, several strategies have been proposed in broader robotic contexts to bridge this reality gap. Data-driven co-design approaches \cite{Schaff_2} have relied on Domain Randomisation \cite{Tan}, while more conventional optimal control strategies have focused on feedforward \cite{Ha} or feedback sensitivity analysis \cite{Giordano}, as well as Stochastic Programming \cite{Palacios}\cite{Palacios_2}. Notably, a bi-level robust co-design framework was proposed for legged robots using an ensemble of stochastic simulations to couple hardware parameters with stabilising linear controllers \cite{Fadini_2}. Such an approach is particularly compelling for agile flight systems, where robustness is critical and co-design strategies often drive the system toward morphologies with inherent instability to maximise manoeuvrability. In these cases, nominally optimal solutions become sensitive to model uncertainty and environmental disturbances. This remains a significant limitation in recent fixed-wing UAV co-design frameworks, which prioritise time optimality over robust performance. For instance, while some frameworks are restricted to open-loop control only \cite{Bergonti}, others co-optimise feedback gains but only validate the final design's robustness against wind disturbances post-optimisation \cite{Zhao}, rather than treating robustness as a primary objective within the loop.

In this work, we propose a framework for robust co-design optimisation of agile fixed-wing UAVs. An overview of the method is illustrated in Fig. \ref{overview}. Central to this approach is a nonlinear aerodynamic model that captures the dynamics across the full flight envelope required for analysing aggressive manoeuvres. Our bi-level co-design strategy consists of a high-level evolutionary design optimisation and a nested low-level evaluation loop that performs optimal trajectory planning and feedback stabilisation. We use a Linear Quadratic Regulator (LQR) for state feedback control under uncertainty due to its high computational efficiency, stability, and robustness properties \cite{anderson2007optimal}. The nested low-level loop decouples the trajectory optimisation from the feedback control task to handle hard state and actuator constraints more robustly in spatially-constrained tasks, and avoid numerical instabilities and discontinuities in the objective landscape. 

Drawing inspiration from Domain Randomisation, our method utilises a nominal system model to identify optimal trajectories and control gains, subsequently incorporating stochastic variations through model mismatch and external disturbances. Robustness is embedded directly into the co-design optimisation via a Monte Carlo ensemble of perturbed simulations for each design and trajectory configuration, with performance quantified by the aggregate cost. This approach ensures the discovery of robust, jointly optimal solutions within non-convex landscapes, effectively rejecting parametric uncertainty and external noise during the optimisation process. 
We benchmark the proposed framework across three agile flight tasks evaluated against a weighted time-energy cost function: (i) an obstacle avoidance mission adapted from \cite{Bergonti}, (ii) a vertical reversal task, and (iii) a spatially-constrained 180-degree hairpin.

The paper is structured as follows. In Section \ref{background}, we define the nonlinear aerodynamic model and equations of motion of the fixed-wing UAV, the trajectory optimisation, and the state feedback controller. Section \ref{codesign_framework} introduces the robust co-design framework, including the high-level design search, the low-level evaluation loop, and the general optimisation formulation. In Section \ref{results}, we present the results of the robust co-design strategy compared against a deterministic baseline co-design framework.

\section{BACKGROUND}
\label{background}

This section presents the theoretical background for the proposed co-design optimisation framework. First, we define the UAV platform and the design parameters to be optimised, which characterise
the 6-DoF dynamical system, the underlying nonlinear aerodynamic model, and the feedback controller. Subsequently, we formulate the trajectory optimisation task for generating feasible manoeuvres, followed by the LQR state feedback control architecture to stabilise the UAV under uncertainty. 

\subsection{Platform Definition}

The platform analysed is a fixed-wing UAV with a conventional wing-fuselage-tail architecture, controlled via an electric motor and propeller and standard flight control surfaces (ailerons, elevator and rudder). The UAV is defined by the set of geometric parameters $\mathbf{d} = \{b_w, c_w, x_w\}$, representing the span, chord, and horizontal position of the wing with respect to the fuselage nose. 

\subsection{UAV Modelling}
\label{sec:drone_modelling}

We model the fixed-wing UAV as a 6-DoF rigid body and represent the attitude using unit quaternions to ensure a singularity-free mapping across the full flight envelope \cite{Sobolic}\cite{Borodacz}. To describe the vehicle's motion, we define an inertial, Earth-fixed frame $\mathcal{I}$ and a body-fixed frame $\mathcal{B}$ attached to the UAV's centre of mass. The full state of the aircraft is defined by the vector $\mathbf{x} = [\mathbf{p}^\top, \mathbf{q}^\top, \mathbf{v}_b^\top, \boldsymbol{\omega}_b^\top]^\top \in \mathbb{R}^{13}$. The position $\mathbf{p} = [x_{e}, y_{e}, z_{e}]^\top \in \mathbb{R}^3$ and orientation $\mathbf{q} = [q_{0}, q_{1}, q_{2}, q_{3}]^\top \in \mathbb{S}^3 \subset \mathbb{R}^4$ are defined relative to the inertial frame $\mathcal{I}$. The motion is described by the velocity vector  $\boldsymbol{\nu} = [\mathbf{v}_b^\top, \boldsymbol{\omega}_b^\top]^\top \in \mathbb{R}^6$ expressed in the body frame $\mathcal{B}$, where $\mathbf{v}_b = [u, v, w]^\top$ and $\boldsymbol{\omega}_b = [p, q, r]^\top$ are the translational and angular velocities. The coupling between the body-fixed velocities and the inertial position and orientation is established through the rigid-body kinematic transformations:
\begin{equation}
\label{kinematics1}
    \dot{\mathbf{p}} = \mathbf{R}(\mathbf{q})^\top \mathbf{v}_b , \quad 
    \dot{\mathbf{q}} = \frac{1}{2} \mathbf{q} \otimes \hat{\boldsymbol{\omega}}_b \,,
\end{equation}

\noindent where $\mathbf{R}(\mathbf{q}) \in SO(3)$ is the rotation matrix from $\mathcal{I}$ to $\mathcal{B}$, $\otimes$ denotes the Hamilton product, and $\hat{\boldsymbol{\omega}}_b = [0, \boldsymbol{\omega}_b^\top]^\top$. The translational and rotational equations of motion can be expressed in reduced matrix form as:
\begin{equation}
\label{eqofmotion}
    \mathbf{M} \dot{\boldsymbol{\nu}} + \mathbf{C}(\boldsymbol{\nu})\boldsymbol{\nu} = \mathbf{G}(\mathbf{q}) + \mathbf{F}_{ext}\, ,
\end{equation}
\noindent where:
\begin{itemize}
    \item $\mathbf{M} = \text{diag}(m\mathbf{I}_3, \mathbf{I}_{body})$ is the generalised inertia matrix.
    \item $\mathbf{C}(\boldsymbol{\nu}) = \text{diag}(m [\boldsymbol{\omega}_{b} \times], -[\mathbf{I}_{body} \boldsymbol{\omega}_{b} \times])$ is the Coriolis and centrifugal matrix, 
    \item $\mathbf{G}(\mathbf{q})$ is the gravity wrench in the body frame $\mathcal{B}$,
    \item $\mathbf{F}_{ext}$ is the generalised external wrench, the sum of aerodynamic loads $\mathbf{F}_{aero}$ and propulsive forces $\mathbf{F}_{prop}$.
\end{itemize}

To accurately model the aerodynamics of agile flight, we use a Strip Theory-based nonlinear aerodynamic model. The model accounts for aerodynamic stall and post-stall covering the entire flight envelope. Each lifting surface is discretised into $N_{s}$ spanwise strips and the set of 2D aerodynamic coefficients at the $i$-th strip $\mathbf{c}_{a, i} = (C_{L, i}, C_{Di, i}, C_{M0, i})$ is evaluated using a semi-empirical flat-plate model \cite{Khan}:   
\begin{equation}
\label{eq:blending}
\mathbf{c}_{a, i} = \sigma(\alpha) \cdot \mathbf{c}_{att, i} + (1 - \sigma(\alpha)) \cdot \mathbf{c}_{ps, i}\, ,
\end{equation}
where $\sigma(\alpha)$ is a sigmoid blending function defined as:
\begin{equation}
\sigma(\alpha) = \frac{1}{2} \left[ 1 + \tanh \left( k (\alpha_{s} - |\alpha|) \right) \right].
\end{equation}
Here, $\alpha$ is the angle of attack, $k$ is the sharpness parameter, and $\alpha_{s}$ is the stall angle. The sigmoid blending function is integrated to ensure $C^{\infty}$ continuity between the attached flow regime described by the aerodynamic coefficients $\mathbf{c}_{att, i}$ and the post-stall flow regime defined by $\mathbf{c}_{ps, i}$. Potential flow theory is used pre-stall, and Hoerner's empirical model is used in the post-stall regime \cite{Hoerner}. The model provides a differentiable representation of the aerodynamic forces across the full flight envelope, required by the gradient-based optimisation approach. Finally, we incorporate empirical 3D corrections based on the wing Aspect Ratio ($AR$).

To estimate the aerodynamic loads over each lifting surface, the sectional forces $f_{i}$ and moments $m_{i}$ are numerically integrated across the $N_s$ spanwise strips. At the $i$-th strip, the loads are recovered from the coefficients in Eq.~\eqref{eq:blending}:  
\begin{equation}
\label{eq:sectional_loads}
\mathbf{f}_{i} = \frac{1}{2}\rho V_{\infty,i}^2 c_i \begin{bmatrix} C_{D,i} \\ 0 \\ C_{L,i} \end{bmatrix}, \quad 
\mathbf{m}_{i} = \frac{1}{2}\rho V_{\infty,i}^2 c_i^2 \begin{bmatrix} 0  \\ C_{M0,i} \\ 0 \end{bmatrix},
\end{equation}
where $C_{D,i}$ includes both the parasitic drag $C_{D0}$ and the induced drag $C_{Di, i}$ calculated from the flat-plate model \cite{Khan}. For vertical surfaces, the normal component is mapped to the lateral axis as $\mathbf{f}_i = [D_i, L_i, 0]^\top$. We assume spanwise independence, neglecting cross-flow effects and aerodynamic coupling between adjacent strips. Furthermore, the fuselage is treated as a non-lifting body, neglecting its aerodynamic contribution and interference with the wing and tail components. With these assumptions, the total aerodynamic wrench $\mathbf{F}_{aero}$ is the sum of the aerodynamic loads on each lifting surface (wing, horizontal and vertical tail) in the set $\{S\}$, expressed in the body frame $\mathcal{B}$ as:

\begin{equation}
\label{eq:total_wrench}
\mathbf{F}_{aero} = \sum_{j \in \{S\}} \sum_{i=1}^{N_s} w_i \begin{bmatrix} \mathbf{R}_i \mathbf{f}_{i} \\ \mathbf{r}_i \times (\mathbf{R}_i \mathbf{f}_{i}) + \mathbf{R}_i \mathbf{m}_{i}, \end{bmatrix},
\end{equation}
where $w_i$ is the spanwise integration weight, $\mathbf{r}_i$ is the position vector from the centre of gravity and $\mathbf{R}_i$ is the rotation matrix from $\mathcal{I}$ to $\mathcal{B}$ at the $i$-th strip. The control surface deflections $\boldsymbol{\delta}_{i} = [\delta_{e}, \delta_{a}, \delta_{r}]^\top$ are modeled as local geometric perturbations to the strips and modify the sectional zero-lift angle and effective camber. This formulation captures the coupling between control authority and the non-uniform dynamic pressure distribution across the span, particularly during high-rate rolling or yawing manoeuvres. We assume quasi-steady aerodynamics, but apply first-order kinematic constraints on the actuator dynamics to account for the finite bandwidth of the hardware: $|\dot{\delta}_i| \le \dot{\delta}_{max}$. The propulsive forces $\mathbf{F}_{prop}$ are represented by the thrust input $T$, positive along the longitudinal axis, assuming instantaneous thrust response and no interaction with the flow field.

\subsection{Trajectory Optimisation}

The agile flight trajectory optimisation task is formulated as a multi-objective Optimal Control Problem (OCP) to minimise a weighted cost of mission duration and energy consumption, defined in Eq. \eqref{cost}. The OCP seeks the optimal sequence of states $\mathbf{x}(t) \in \mathbb{R}^{13}$ and control inputs $\mathbf{u}(t) \in \mathbb{R}^{4}$ to fly the UAV along the most efficient trajectory over the time interval $t \in [t_{0}, t_{f}]$, where $\mathbf{u} = [T, \boldsymbol{\delta^{\top}}]^{\top}$ comprises the thrust $T$ and the control surface deflection vector $\boldsymbol{\delta}$. To solve the OCP, the trajectory optimisation is transcribed into a finite-dimensional Nonlinear Programming (NLP) problem via Direct Collocation and discretised into $N$ knots. Rather than integrating the dynamics sequentially, Direct Collocation treats the states $\mathbf{x}_{k}$ and control inputs $\mathbf{u}_{k}$ at each knot $k = \{0, 1,\dots, N - 1\}$ as independent decision variables, along with the terminal time $t_{f}$ which dictates the uniform time step $\Delta t = t_{f} / (N - 1)$. The physical validity of the trajectory is ensured by constraining the states and control inputs to satisfy the system dynamics $\dot{\mathbf{x}} = \mathbf{f}(\mathbf{x}, \mathbf{u})$ at the collocation points. This formulation yields a large-scale but sparse NLP that improves the convergence rate, prevents numerical divergence and facilitates the embedding of path constraints. The trajectory optimisation problem is defined as:
\begin{align}
\underset{\mathbf{x}_{0 \dots N}, \mathbf{u}_{0 \dots N}, t_f}{\text{min}} \quad & \mathcal{L}(\mathbf{x}, \mathbf{u}) \label{trajopt} \\
\text{s.t. } \quad & \boldsymbol{\zeta}_k = \mathbf{0}, && \text{Eq. \eqref{dynamics}} \nonumber \\
& \mathbf{g} = \mathbf{0}, && \text{Eq. \eqref{initial}, \eqref{terminal}} \nonumber \\
& \mathbf{h} \leq \mathbf{0}, && \text{Eq. \eqref{hardware1} - \eqref{obstacles}} \nonumber \
\end{align} where $\boldsymbol{\zeta}_k(\mathbf{x}_k, \mathbf{x}_{k+1}, \mathbf{u}_k, \mathbf{u}_{k+1})$ are the dynamic constraints, $\mathbf{h}(\mathbf{x}_k, \mathbf{u}_k)$ are path constraints, and $\mathbf{g}(\mathbf{x}_0, \mathbf{x}_N)$ are boundary constraints. The cost function $\mathcal{L}(\mathbf{x}, \mathbf{u})$ aggregates the mission time and energy efficiency components weighted by coefficients $\omega_1$ and $\omega_2$:
\begin{equation}
\label{cost}
\mathcal{L}(\mathbf{x}, \mathbf{u}) = \omega_1 t_{f} + \omega_2 \sum_{k=1}^{N} \left( \frac{T_{k} \cdot V_{k}}{\eta_{p}} + k_{s} \|\boldsymbol{\delta_{k}}\|^2 \right) \Delta t\, ,
\end{equation}
where $T_{k}$ and $V_{k}$ are the thrust input and airspeed at knot $k$, while $\eta_{p}$ and $k_{s}$ represent the propeller efficiency and the power-to-deflection coefficient for the control inputs $\boldsymbol{\delta_{k}}$, respectively. The system dynamics $\dot{\mathbf{x}} = \mathbf{f}(\mathbf{x}, \mathbf{u})$ are derived by coupling the rigid-body kinematics in Eq. \eqref{kinematics1} with the equations of motion in Eq. \eqref{eqofmotion}. The dynamics are enforced as equality constraints at each knot $k$, using the implicit trapezoidal rule for collocation: 
\begin{equation}
\label{dynamics}
    \boldsymbol{\zeta}_{k} = \mathbf{x}_{k+1} - \mathbf{x}_{k} - \frac{\Delta t}{2} (\mathbf{f}_{k} + \mathbf{f}_{k+1}) = \mathbf{0}.
\end{equation}

The optimisation is subject to physical and operational constraints that arise from the nature of the task or the hardware limitations of the fixed-wing UAV platform:
\subsubsection{Initial Conditions} enforced on all states at the first knot, $k$ = 0:
\begin{equation}
\label{initial}
\mathbf{x}(t_0) = \mathbf{x}_0.
\end{equation}

\subsubsection{Terminal Conditions} enforced on position, orientation and linear velocity at the last knot, $k$ = $N$:
\begin{equation}
\label{terminal}
    \mathbf{p}(t_{N}) =  \mathbf{p}_N, \ 
    \mathbf{q}(t_{N}) =  \mathbf{q}_N, \
    \mathbf{v}_{b}(t_{N}) =  \mathbf{v}_{b,N} .
\end{equation}

\subsubsection{Hardware Limitations} enforced on the thrust input and control surface deflections at each knot $k$, to replicate hardware saturation. First-order kinematic constraints are enforced as explained in Section \ref{sec:drone_modelling}:
\begin{equation}
\label{hardware1}
    T_{min} \leq T(t_{k}) \leq T_{max}, \ \boldsymbol{\delta}_{min} \leq \boldsymbol{\delta}(t_{k}) \leq \boldsymbol{\delta}_{max} ,
\end{equation}
\begin{equation}
\label{hardware2}
    \dot{\boldsymbol{\delta}}_{min} \leq \dot{\boldsymbol{\delta}}(t_{k}) \leq \dot{\boldsymbol{\delta}}_{max} .
\end{equation}

\subsubsection{Quaternion Unit} relaxed inequality constraint enforced on the quaternion norm at each knot $k$ to maintain the attitude on the $SO(3)$ manifold with a tolerance $\epsilon_{tol}$:
\begin{equation}
\label{quaternion}
    1 - \epsilon_{tol} \leq \|\mathbf{q}_{k}\|_{2}^{2} \leq 1 + \epsilon_{tol} .
\end{equation}

\subsubsection{Obstacle Avoidance} no collision, mission-specific constraints enforced at each knot $k$ for all cylindrical obstacles $j \in \mathcal{O}$, with radius $R$ and centred at $\mathbf{p}_{c,j} = (x_{c,j}, y_{c,j}, z_{c,j})$:
\begin{equation}
\label{obstacles}
    \| \mathbf{A}(\mathbf{p}_{k} - \mathbf{p}_{c,j}) \|_{2} \geq R + 0.5 b_{w} ,
\end{equation}
where $\mathbf{A} = \text{diag}(1, 1, 0)$ is an orthogonal projection matrix.

\subsection{State Feedback Control}

To consider parametric uncertainties and external disturbances, a UAV state feedback controller is implemented to track the optimal trajectory generated by the OCP. While deploying an open-loop trajectory optimisation in co-design tasks can discover an optimal configuration and a set of performance bounds under nominal conditions \cite{Bergonti}, integrating a feedback controller ensures that the task is executed robustly in the presence of disturbances. Traditional control architectures for autonomous fixed-wing UAVs rely on cascaded controllers with successive loop closure \cite{Bulka}, that usually decouple the longitudinal and lateral dynamics \cite{Roland}. However, for agile flight manoeuvres, longitudinal and lateral dynamics are significantly coupled. Model Predictive Control (MPC) has been studied for agile flight control \cite{Reinhardt}, but solving a constrained optimisation problem at every time step is computationally expensive within a robust co-design framework. However, if we linearise the dynamics along the trajectory, MPC is equivalent to a local linear controller \cite{Fadini_2}. This motivates the use of an adapted Time-Varying LQR controller \cite{Tedrake} which performs online successive linearisation of the system dynamics $\dot{\mathbf{x}_t} = \mathbf{f}(\mathbf{x}_t, \mathbf{u}_{t}) \approx \mathbf{A}_t \mathbf{x}_t + \mathbf{B}_t \mathbf{u}_{t}$ about the actual state $\mathbf{x}_t$ and control input $\mathbf{u}_{t}$, where $\mathbf{A}_t$ and $\mathbf{B}_t$ are the linearised state-space matrices:
\begin{equation}
\mathbf{A}_t = \frac{\partial \mathbf{f}}{\partial \mathbf{x}} \bigg|_{\mathbf{x}_t, \mathbf{u}_{t}}, \quad \mathbf{B}_t = \frac{\partial \mathbf{f}}{\partial \mathbf{u}} \bigg|_{\mathbf{x}_t, \mathbf{u}_{t}}.
\end{equation}

The main advantage of our approach over standard LQR with time-indexed gain scheduling lies in the controller's ability to maintain local optimality even under significant deviations from the reference trajectory. For our application, this can improve the robustness to large wind gusts and nonlinearities induced by aerodynamic stall. To find the feedback gain matrix $\mathbf{K}_t$ at time $t$, we solve the Algebraic Riccati Equation that guarantees a locally optimal solution at the linearisation point. The control law aggregates the feedforward reference control input $\mathbf{u}^{*}_t$ with a state feedback term that steers the system back to the reference state $\mathbf{x}^{*}_t$:
\begin{equation}
    \mathbf{u}_{t} = \mathbf{u}^{*}_{t} - \mathbf{K}_{t} (\mathbf{x}_{t} - \mathbf{x}^{*}_{t}).
\end{equation}

Due to the use of unit quaternions for a singularity-free attitude representation, the global state $\mathbf{x}_t \in \mathbb{R}^{13}$ is mapped to an error vector $\tilde{\mathbf{x}}_t = \mathbf{x}_t - \mathbf{x}^{*}_t \in \mathbb{R}^{12}$ to operate on a minimal coordinate set. While position and velocity errors use element-wise subtraction, the attitude error $\tilde{\boldsymbol{\theta}}_{t} \in \mathbb{R}^3$ is defined through a multiplicative formulation that maps the manifold $SO(3)$ to its tangent space. Using small-angle approximation within this linear regime, the attitude error is:
\begin{equation}
        \tilde{\boldsymbol{\theta}}_{t} = 2 \cdot \text{sgn}(q_{w,t}) \cdot \mathbf{q}_{v,t} , \quad \tilde{\mathbf{q}}_{t} = \mathbf{q}^*_t \otimes \mathbf{q}_t = \begin{bmatrix} q_{w,t} \\ \mathbf{q}_{v,t} \end{bmatrix},
\end{equation}
where $\mathbf{q}^*_t$ is the quaternion conjugate.

\section{ROBUST CO-DESIGN FRAMEWORK}
\label{codesign_framework}

We propose a bi-level co-design framework to optimise a fixed-wing UAV under uncertainty. Robustness is assessed by subjecting each candidate design to an ensemble of stochastic simulations. The ensemble exposes the model to both parametric uncertainty, such as manufacturing-induced variability in the design parameters, as well as to stochastic wind disturbances. The bi-level structure splits the problem into a high-level design search loop that evolves the hardware parameters, and a low-level evaluation loop that quantifies the performance of each candidate design, as shown in Fig. \ref{overview}. 

\subsection{High-Level Design Search}

To navigate the non-convex and potentially non-smooth cost landscape, the outer design search loop is performed using the Covariance Matrix Adaptation Evolution Strategy (CMA-ES) \cite{Hansen}. CMA-ES is derivative-free and can handle stochasticity in the cost function very efficiently by using rank-based selection in the parameter evolution. The method maintains an internal covariance matrix that captures the underlying coupling between design parameters, particularly useful for discovering aero-inertial trade-offs when co-designing agile UAVs. The optimisation is initialised with a population of $N_{pop}$ candidate designs $\mathbf{d} \in \mathcal{D} \subset \mathbb{R}^3 $ sampled from a multivariate Gaussian distribution $\mathbf{d} \sim \mathcal{N}(\mathbf{d}_0, \sigma_0^2 \mathbf{I})$, where $\mathbf{d}_0$ is the baseline configuration and $\sigma_0$ is the initial step size. The search space $\mathcal{D}$ is restricted to a physically realisable regime throughout the optimisation, such that for each candidate design, $\mathbf{d}_{min} \leq \mathbf{d} \leq \mathbf{d}_{max}$. The high-level optimiser feeds the candidate design $\mathbf{d}$ to a low-level evaluation loop, which quantifies the expected performance across the ensemble of $N_{sim}$ stochastic simulations and returns it to CMA-ES as the global cost function $\mathcal{L}_{\boldsymbol{\xi}}$. Given the set of stochastic parameters $\boldsymbol{\xi}_{i}$ for each simulation $i \in \{1,\dots, N_{sim}\}$, the global cost is defined as the aggregate of individual simulation costs $\mathcal{L}(\mathbf{x}_{\boldsymbol{\xi}_i}, \mathbf{u}_{\boldsymbol{\xi}_i})$:
\begin{equation}
\label{cost_global}
    \mathcal{L}_{\boldsymbol{\xi}} = \mathbb{E} [\mathcal{L}_{\boldsymbol{\xi}_i}] = \frac{1}{N_{\text{sim}}} \sum_{i=1}^{N_{\text{sim}}} \mathcal{L}(\mathbf{x}_{\boldsymbol{\xi}_i}, \mathbf{u}_{\boldsymbol{\xi}_i}).
\end{equation}  

The design candidates are sampled from a multidimensional Gaussian distribution, which updates the mean and covariance matrix based on the performance of the most successful candidates, guiding the design search towards the cost-minimising solution. Hence, the robust global optimisation performed by CMA-ES seeks the jointly optimal set of design parameters $\mathbf{d}^{*}$ and controlled trajectory $(\mathbf{x}^{*}, \mathbf{u}^{*})$ that satisfy: 

\begin{equation}
\mathbf{d}^{*}, \mathbf{x}^{*}, \mathbf{u}^{*} = \underset{\mathbf{d}, \mathbf{x}, \mathbf{u}}{\text{argmin}} \quad \mathcal{L}_{\boldsymbol{\xi}}.
\end{equation} 

We perform the optimisation for $N_{gen}$ generations or until cost $\mathcal{L}_{\boldsymbol{\xi}}$ and design parameters $\mathbf{d}$ converge to $|\mathcal{L}_{\boldsymbol{\xi}} - \mathcal{L}_{\boldsymbol{\xi}_{prev}}| < tol_{c}$ and $|\mathbf{d} - \mathbf{d}_{prev}| < tol_{p}$. To speed up the co-design optimisation, we execute CMA-ES in parallel and evaluate candidate designs concurrently.

\begin{table*}[ht]
\centering
\caption{Comparison of nominal and robust co-design optimisation results across the three agile flight tasks. The post-optimisation robustness analysis evaluates each robust optimal design against the nominal design under the same parameter uncertainty and wind disturbances. The UAV schematics and target gates (green circle) are scaled by a factor of five.}
\footnotesize
\renewcommand{\arraystretch}{1.2}
\newcolumntype{Y}{>{\centering\arraybackslash}X} 
\begin{tabularx}{\textwidth}{@{} c c YYY YY YY @{}}
\toprule
\multirow{2}{*}{\textbf{Task}} & \multirow{2}{*}{\textbf{Optimisation Scenario}} & \multicolumn{3}{c}{\textbf{Optimal Design}} & \multicolumn{2}{c}{\textbf{Eval. RMSE (m)}} & \multicolumn{2}{c}{\textbf{Eval. Success Rate}} \\
\cmidrule(lr){3-5} \cmidrule(lr){6-7} \cmidrule(lr){8-9}
& & $b_{w}\, (\si{m})$ & $c_{w}\, (\si{m})$ & $x_{w}\, (\si{m})$ & Nominal & Robust & Nominal & Robust \\ \midrule

\multirow{6}{*}{\shortstack{\textbf{Obstacle Avoidance} \\ \includegraphics[width=4.1cm, valign=c]{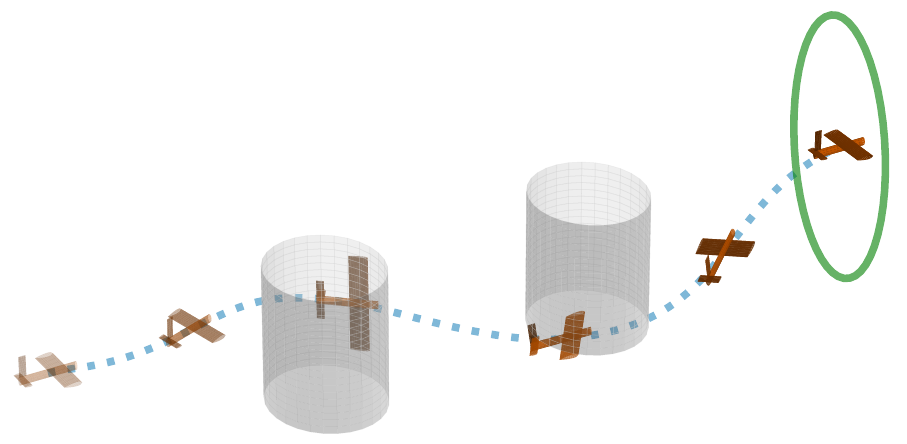}}}
& Nominal        & 1.000 & 0.116 & 0.054 & - & - & - & - \\
& 5\% UC         & 0.998 & 0.131 & 0.106 & 25.132 & \textbf{0.243} & 0\% & \textbf{100\%} \\
& 10\% UC        & 0.998 & 0.165 & 0.133 & 24.372 & \textbf{0.243} & 0\% & \textbf{100\%} \\
& Wind           & 0.974 & 0.171 & 0.306 & 30.536 & \textbf{0.306} & 0\% & \textbf{100\%} \\
& 5\% UC + Wind  & 1.000 & 0.169 & 0.324 & 29.078 & \textbf{0.286} & 0\% & \textbf{100\%} \\
& 10\% UC + Wind & 0.944 & 0.180 & 0.311 & 29.700 & \textbf{0.302} & 0\% & \textbf{100\%} \\ 
\midrule 

\multirow{6}{*}{\shortstack{\textbf{Vertical Reversal} \\ \includegraphics[width=2.6cm, valign=c]{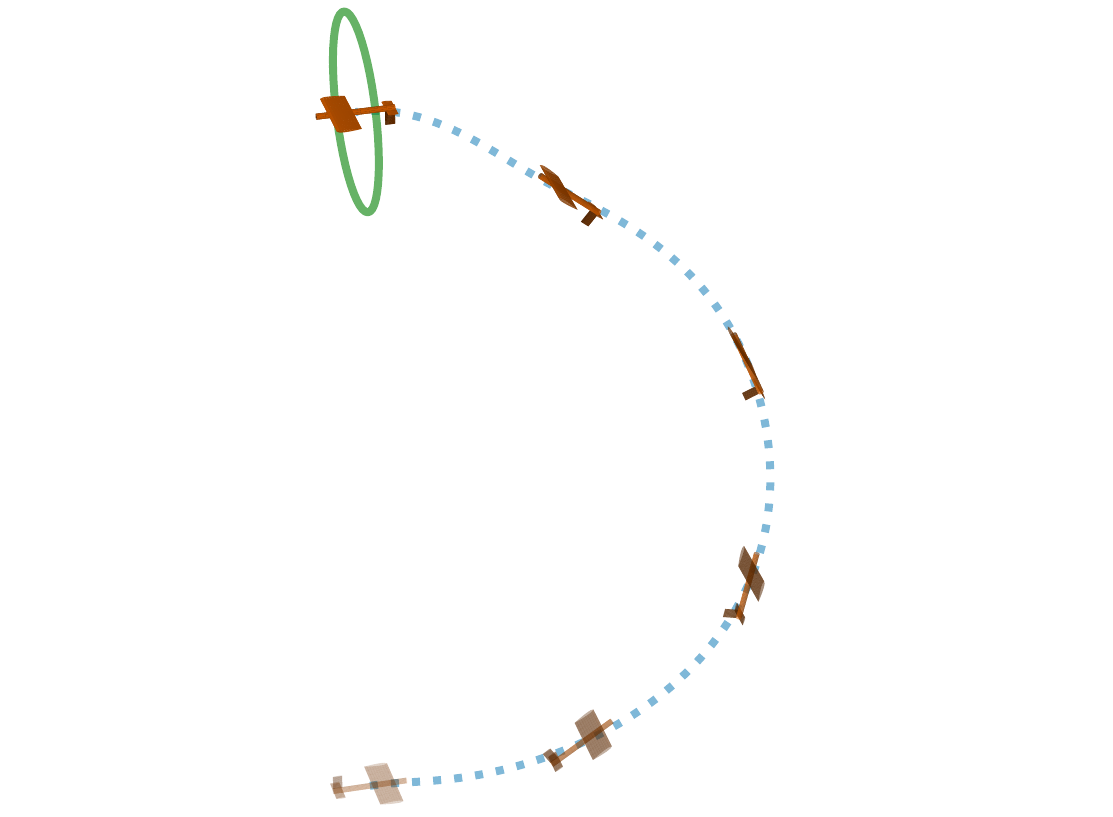}}}
& Nominal        & 1.000 & 0.071 & 0.060 & - & - & - & - \\
& 5\% UC         & 0.976 & 0.103 & 0.194 & 26.374 & \textbf{0.322} & 0\% & \textbf{100\%} \\
& 10\% UC        & 1.000 & 0.126 & 0.226 & 21.866 & \textbf{0.469} & 0\% & \textbf{100\%} \\
& Wind           & 1.000 & 0.133 & 0.325 & 35.465 & \textbf{0.437} & 0\% & \textbf{100\%} \\
& 5\% UC + Wind  & 1.000 & 0.140 & 0.324 & 34.059 & \textbf{0.518} & 0\% & \textbf{100\%} \\
& 10\% UC + Wind & 0.996 & 0.155 & 0.328 & 35.837 & \textbf{0.730} & 0\% & \textbf{99\%} \\ 
\midrule 

\multirow{6}{*}{\shortstack{\textbf{Horizontal Hairpin} \\ \includegraphics[width=3.1cm, valign=c]{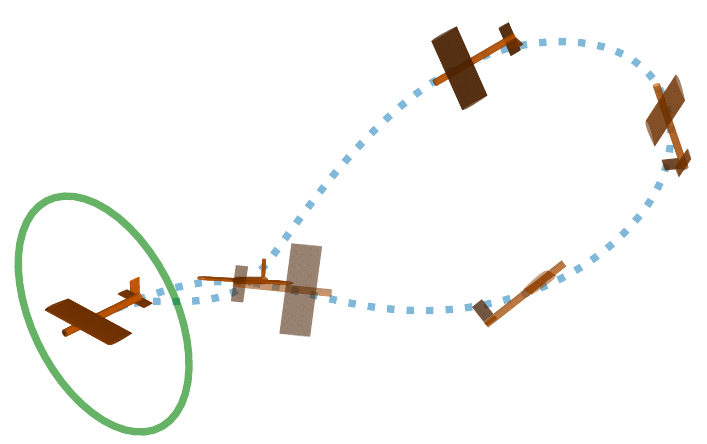}}}
& Nominal        & 0.996 & 0.310 & 0.245 & - & - & - & - \\
& 5\% UC         & 0.982 & 0.300 & 0.295 & 0.303 & \textbf{0.349} & 98\% & \textbf{100\%} \\
& 10\% UC        & 0.948 & 0.316 & 0.340 & 0.670 & \textbf{0.544} & 85\% & \textbf{99\%}  \\
& Wind           & 0.998 & 0.313 & 0.433 & 0.799 & \textbf{0.448} & 69\% & \textbf{100\%}  \\
& 5\% UC + Wind  & 0.662 & 0.298 & 0.318 & 0.901 & \textbf{0.659} & 68\% & \textbf{90\%}  \\
& 10\% UC + Wind & 0.894 & 0.324 & 0.324 & 1.215 & \textbf{0.691} & 54\% & \textbf{85\%}  \\ 
\bottomrule
\end{tabularx}
\label{tab:optimization_results}
\end{table*}

\subsection{Low-Level Evaluation Loop}

The low-level evaluation loop quantifies the task performance of the candidate designs. First, an optimal trajectory is discovered using a nominal model of the system and disturbance-free dynamics. The trajectory is then tracked by a state feedback controller across a stochastic ensemble of simulations to quantify robustness against parametric uncertainties and external disturbances. Each candidate design $\mathbf{d}$ is transformed into the set of nominal physical parameters $\bar{\boldsymbol{\xi}} = \mathcal{M}(\mathbf{d})$ via the mapping function $\mathcal{M}(\cdot)$. This nominal model $\bar{\boldsymbol{\xi}}$ comprises the idealised physical properties of the airframe, which define the system dynamics $\dot{\mathbf{x}} = f(\mathbf{x}, \mathbf{u}, \bar{\boldsymbol{\xi}})$ for the open-loop OCP. This is transcribed into a constrained NLP problem solved using Eq. \eqref{trajopt} to generate the time-optimal reference trajectory $(\mathbf{x}_{n}^{*}, \mathbf{u}_{n}^{*})$ that minimises the nominal cost $\mathcal{L}_{\bar{\boldsymbol{\xi}}}$ defined as in Eq. \eqref{cost}. We use a Monte Carlo approach to evaluate each candidate design across an ensemble of stochastic simulations. In each instance, an adapted Time-Varying LQR state feedback controller stabilises the system about the reference trajectory. The controller uses the same nominal model $\bar{\boldsymbol{\xi}}$ as the open-loop trajectory planning to compute the control gains, whereas each stochastic simulation in the ensemble utilises a perturbed model $\boldsymbol{\xi}_{i}$ in the dynamics. This accounts for parametric discrepancies in the design that arise from variability in manufacturing and assembly processes and affect the mass, inertia, and aerodynamic properties. Each perturbed model $\boldsymbol{\xi}_{i}$ is obtained from a Gaussian distribution modelling the variability of the co-design parameters:

\begin{equation}
    \boldsymbol{\xi}_{i} = \mathcal{M}(\mathbf{d} + \boldsymbol{\epsilon}), \quad \boldsymbol{\epsilon} \sim \mathcal{N}(0, \sigma^{2}) \mathbf{d}. 
\end{equation}
 
To account for external disturbances, we introduce continuous atmospheric turbulence into the simulated dynamics using the stochastic Von K{\'a}rm{\'a}n model in line with airworthiness certification requirements \cite{gust}. The model uses spectral shaping filters to replicate the frequency-dependent energy distribution of turbulence in the velocity field $\mathbf{v}_{w}$. Therefore, the perturbed dynamics are defined by $\dot{\mathbf{x}} = f(\mathbf{x}, \mathbf{u}, \boldsymbol{\xi}_{i}, \mathbf{v}_{w})$. The individual simulation cost function $\mathcal{L}(\mathbf{x}_{\boldsymbol{\xi}_{i}}, \mathbf{u}_{\boldsymbol{\xi}_{i}})$ aggregates the nominal cost $\mathcal{L}_{\bar{\boldsymbol{\xi}}}$ defined in Eq. \eqref{cost} with tracking and mission success metrics. 

The open-loop planning and the LQR feedback control are treated as separate tasks to ensure strict enforcement of state and actuator limits as hard constraints. This guarantees that the reference trajectory is physically feasible before stabilisation and allows the synthesis of control gains directly on the simulated path. The controller maintains constant weight matrices $\mathbf{Q}$ and $\mathbf{R}$ in simulations to ensure an unbiased evaluation across the design candidates.

\section{RESULTS}
\label{results}
We apply our robust co-design optimisation method to three agile flight tasks shown in Table \ref{tab:optimization_results}, including an obstacle avoidance mission, a vertical reversal task, and a horizontal hairpin manoeuvre. We compare our framework against a deterministic co-design baseline \cite{Bergonti}. In the baseline method, the design search is coupled with an open-loop trajectory optimisation under nominal conditions, and does not consider wind disturbances and model parameter uncertainty.

\subsection{Framework Setup}
The high-level design search uses CMA-ES with a population size of $N_{pop} = 32$, and performed over $N_{gen} = 100$ generations for the vertical reversal and obstacle avoidance tasks, and $N_{gen} = 150$ for the horizontal hairpin task. We define early termination criteria based on two convergence thresholds: the cost threshold $tol_{c}=10^{-4}$ and the design parameter threshold $tol_{p}=10^{-4}$. For the nonlinear aerodynamic model, $N_{s} = 6$ spanwise strips per lifting surface allow accurate prediction of local stall and roll-damping during agile manoeuvres, while remaining computationally efficient. The low-level trajectory optimisation is formulated using the \texttt{CasADi} \cite{casadi} framework for automatic differentiation and is solved with \texttt{IPOPT} \cite{ipopt} and the \texttt{MUMPS} \cite{mumps} linear solver. 
The robustness of each design is evaluated using a Monte Carlo ensemble of $N_{sim} = 100$ noisy simulations. 
For the LQR controller, the weight matrices $\mathbf{Q}$ and $\mathbf{R}$ are task-specific but do not require retuning, as the sensitivity to parametric changes of the system dynamics is inherently embedded in the design-dependent linearisation matrices $\mathbf{A}_t$ and $\mathbf{B}_t$. We run our co-design optimisation on an Intel Icelake Xeon Platinum 8358 machine, with the CMA-ES search parallelised across 32 CPU cores. The optimisation takes an average of $\sim27.4 \, \si{h}$ to converge, and evaluates 3200 or 4800 design candidates depending on the task. 

The fixed-wing UAV platform uses conventional control surfaces (ailerons, elevator, and rudder) with a constant chord fraction $c_{f}/c = 0.3$. The cylindrical fuselage has a radius of $0.05\, \si{m}$ and a length of $0.8 \, \si{m}$, considering an additional payload of $0.35 \, \si{kg}$ to account for the avionics and battery, located at the centre of gravity. The conventional tail consists of a horizontal and vertical stabiliser, both with a span of $0.3\, \si{m}$ and chord of $0.1\, \si{m}$. All lifting surfaces use NACA 0012 profiles with a blending sharpness $k = 20$. This approximates the progressive stall characteristics across the operational range $Re = 3.4 \times 10^4 - 3.4 \times 10^5$ without introducing gradient discontinuities in the NLP. The co-design parameters are the wing span $b_{w} \in [0.05, 1]\, \si{m}$, the wing chord $c_{w} \in [0.05, 0.5]\, \si{m}$, and the wing horizontal location  $x_w \in [0.05, 0.5]\, \si{m}$ defined from the nose of the fuselage. These parameters directly influence two primary fixed-wing agility characteristics: the aspect ratio $AR = b_{w} / c_{w}$ that defines the roll moment of inertia and dictates the aerodynamic efficiency and resistance to banking, and the static margin $K_{n} = x_{np}-x_{cg}$ that governs the longitudinal stability and pitch responsiveness. This distance $K_{n}$ between the neutral point (where aerodynamic forces act) and the centre of gravity defines the longitudinal stiffness and indicates if the UAV is naturally stable under pitch disturbances. While traditional static approaches optimise for steady-state manoeuvrability and aerodynamic efficiency, these metrics in isolation do not guarantee that a specific agile flight mission can be performed. Therefore, a dynamic co-design formulation is required to evaluate the vehicle’s performance across the entire manoeuvre manifold rather than at isolated equilibrium points. Hence, our framework ensures that the optimal designs possess the control authority and inertial properties necessary to robustly perform complex agile manoeuvres in spatially-constrained environments.

We evaluate our method on an obstacle avoidance mission, a vertical reversal task, and a constrained hairpin manoeuvre. The initial condition is consistent across the three tasks: the UAV starts from $\mathbf{p}_{0} = (0, 0, 0) \, \si{m}$ with a velocity $\mathbf{v}_{b, 0} = (10, 0, 0) \, \si{m/s}$. The terminal states of the three tasks are: 

\begin{table}[h!]
\centering
\label{tab:terminal_states}
\begin{tabular}{cccc}
\toprule
\textbf{Task} & $\mathbf{p}_{N}$ (m) &  $\mathbf{v}_{b, N}$ (m/s) &  $\mathbf{q}_{N}$ \\ \midrule
Obstacle Avoidance & $(60, 0, 0)$ & $(10, 0, 0)$ & $(1, 0, 0, 0)$ \\
Vertical Reversal & $(0, 0, 40)$ & $(10, 0, 0)$ & $(0, 0, 1, 0)$ \\
Horizontal Hairpin & $(0, 0, 0)$ & $(10, 0, 0)$ & $(0, 0, 0, 1)$ \\ \bottomrule
\end{tabular}
\end{table} 

\noindent This formulation does not pre-define the desired trajectories, but ensures that they emerge as an optimal path between the initial and terminal states to minimise the weighted time-energy cost while satisfying the UAV dynamics. For the obstacle avoidance task, there are two cylinders of infinite length above ground and radius $R = 4 \, \si{m}$ centred at $\mathbf{p}_{c, 1} = (20, 0, 0) \, \si{m}$ and $\mathbf{p}_{c, 2} = (40, 0, 0) \, \si{m}$. Hardware constraints are enforced on the control surface deflections and rates  $\delta_{i} \in [-25^{\circ}, 25^{\circ}]$ and  $\dot{\delta}_{i} \in [-8, 8] \, \si{rad/s}, \ i \in \{e, a, r\} $ \cite{Dorobantu}. The thrust input is bounded to $T \in [0, 5] \, \si{N}$ for the obstacle avoidance task and $T \in [0, 10] \, \si{N}$ for the hairpin and vertical reversal tasks. The cost function uses $\omega_{1} = 1.0$ and $\omega_{2} = 0.1$ as weighting coefficients to maintain a trade-off between time optimality and energy consumption at a comparable order of magnitude. The propeller efficiency is set to $\eta_{p} = 0.65$, conservative for small-scale DC motors, while the power-to-deflection coefficient is $k_{s} = 0.5$ to avoid excessive control authority usage. To quantify robustness within the co-design optimisation, we use two metrics: the position tracking $\text{RMSE} = \sqrt{ \sum_{i=1}^{N_{sim}} \left\| \mathbf{p} - \mathbf{p}^{*} \right\|^2 / N_{sim}} $ and the success rate $\eta = \sum_{i=1}^{N_{sim}} S_{i} / N_{sim}$ across the Monte Carlo ensemble. Success is defined by the binary variable $S_{i}$, which equals one if the UAV avoids all obstacles (satisfying Eq. \eqref{obstacles}) and terminates within a gate of radius $R_{g} = 1 \,  \si{m}$.

\subsection{Co-design Results}
We compare our robust co-design optimisation frameworks with a deterministic co-design baseline in which each UAV design is evaluated exclusively by an open-loop trajectory optimisation task under nominal conditions. We call this \textbf{Nominal Co-design}, where the overall cost $\mathcal{L}_{\bar{\boldsymbol{\xi}}}$ is only a function of the candidate design $\mathbf{d}$ and the optimal trajectory $(\mathbf{x}_{n}^{*}, \mathbf{u}_{n}^{*})$. For the \textbf{Robust Co-design} framework, we run each task across five perturbed scenarios: 5\% and 10\% uncertainty in the design parameters, wind-only disturbances, and a combined case with both wind disturbances and parametric uncertainty at the two specified intensities. Our framework evaluates design candidates based on the cost function $\mathcal{L}_{\boldsymbol{\xi}}$, formulated in Eq. \eqref{cost_global}. To compare the nominal and robust designs, we perform a post-optimisation robustness analysis by evaluating the optimal solutions over an ensemble of noise realisations within a stochastic simulation framework. By computing the RMSE and success rate across the ensemble, we quantify the robustness of the nominal and robust designs under parametric uncertainty and environmental disturbances.

An overview of the designs optimised across all tasks and scenarios is presented in Table \ref{tab:optimization_results}. In the obstacle avoidance and vertical reversal tasks, all nominal and robust UAV configurations maximise the aspect ratio $AR$, mainly by pushing the wing span $b_{w}$ to the upper bound of the design search space. By weighting energy consumption more heavily than mission time, the optimisation favors high-$AR$ designs that minimise induced drag. This design choice maximises aerodynamic efficiency, deliberately trading a slight increase in mission time for improved energy performance.

\begin{figure*}[t] 
    \centering
    \includegraphics[width=1.0\textwidth]{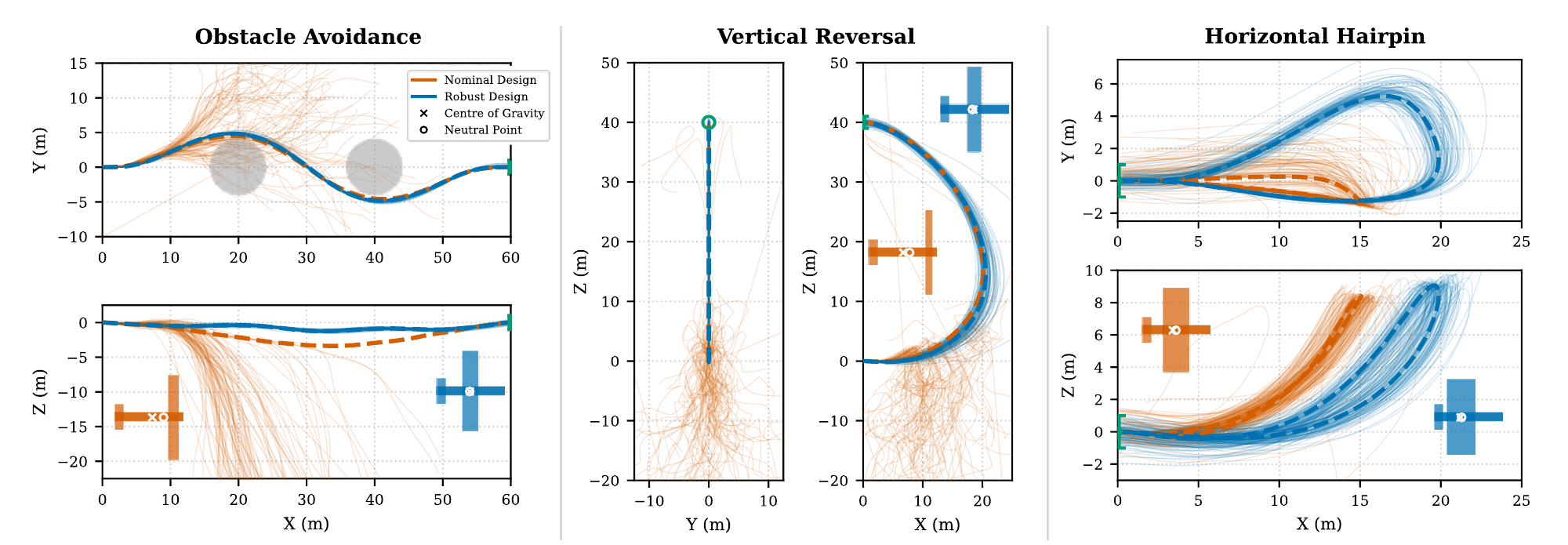}
    \caption{Robustness analysis performed on nominal (orange) and robust (blue) optimal designs under 10\% parametric noise and wind disturbances. Faint traces represent individual trajectories, dashed bold lines show the reference trajectories. The centre of gravity $x_{cg}$ and neutral point $x_{np}$ are highlighted on each design to visualise the static margin $K_{n}$.}
    \label{comparison}
\end{figure*}

For the obstacle avoidance task, we can see a significant shift in the wing longitudinal position $x_{w}$ from $0.054 \, \si{m}$ to $0.311 \, \si{m}$ between the nominal and the highest-uncertainty robust scenario (10\% parametric uncertainty and gust disturbances), in conjunction with a $55\%$ increase in chord size $c_{w}$. In the nominal case, the forward wing placement and negative static margin ($K_{n} = -0.132\, \si{m}$) yields an inherently unstable design that prioritises agility over robustness to maximise mission performance. However, the design is practically uncontrollable under disturbances, indicated by the high RMSE values in Table \ref{tab:optimization_results} and the rapid trajectory divergence across the ensemble of stochastic evaluations in Fig. \ref{comparison}. Moving the wing aft shifts the static margin to a positive, near-neutral regime ($K_{n} = 0.002\, \si{m}$). This yields a robust optimal design that is naturally stable while maintaining the agility required for the aggressive manoeuvring. Rather than relying on a high static margin for stiff passive stability, the design uses an increased chord to provide the necessary damping. This is supported by the significant reduction in the variance of the thrust input and actuator deflections throughout the trajectory: we observe an order of magnitude decrease in $T_{\sigma}$ and $\boldsymbol{\delta}_{\sigma}$ when transitioning from the saturated 'bang-bang' control of the nominal design to the smooth actuations of the robust controller. This confirms that by addressing uncertainty at the design level, the robust co-design ensures high tracking fidelity and a $100\%$ success rate across all tested scenarios. In comparison, the nominal design that relies on aggressive control experiences significant performance degradation under uncertainty and fails to clear obstacles or reach the terminal gate in every instance.

For the vertical reversal task, we can see similar trends in the nominal and robust co-design solutions: the wing position $x_{w}$ moves aft from $0.060 \, \si{m}$ to $0.328 \, \si{m}$ with increasing uncertainty in the system, shifting the static margin from $K_{n} = -0.076\, \si{m}$ to $K_{n} = 0.012\, \si{m}$ and increasing the wing chord from $0.071\, \si{m}$ to $0.155\, \si{m}$. The nominal design has a smaller chord, close to the lower limit of the design space, to limit pitch damping and induced drag, and achieve the aggressive pitch rates required for the 180$^{\circ}$ reversal manoeuvre. However, as shown in Fig. \ref{comparison}, it diverges early from the reference trajectory, failing to reach the target gate and resulting in a 0\% success rate in all scenarios. In contrast, the robust design successfully performs the pull-up manoeuvre, reaching the target gate in 100\% of evaluations for most cases, and maintaining a 99\% success rate even under the most severe disturbance and uncertainty conditions. 

For the horizontal hairpin task, the requirement for a 180$^{\circ}$ yaw reversal drives the optimisation toward lateral agility rather than aerodynamic efficiency. This results in much lower aspect ratios ($AR \approx 3$) for both designs compared to the other two tasks ($AR \approx 6-10$). These UAV designs minimise rotational inertia and reduce roll and yaw damping to improve manoeuvrability for aggressive turning. To compensate, the robust design shifts the wing position aft to $0.324\, \si{m}$ (from $0.245\, \si{m}$ in the nominal case), effectively increasing the static margin and longitudinal stiffness to maintain control under uncertainty. While this reduces the aggressiveness of the highly unstable nominal configuration ($K_{n} = -0.037\, \si{m}$), the robust design deliberately maintains a negative static margin ($K_{n} = -0.013\, \si{m}$ or $-3.9\%$). This induces a degree of longitudinal instability, ensuring the UAV remains agile enough for rapid manoeuvring. The difference between the nominal and robust optimal designs is less pronounced here, as the more extreme mission dynamics enforce more conservative trajectories. While the nominal configuration remains controllable, it exhibits poor performance under uncertainty, yielding only a 54\% success rate and a tracking error of $1.215\, \si{m}$ in the most severe disturbance scenario, shown in Fig. \ref{comparison}. In contrast, the robust co-design framework consistently achieves near $100\%$ success rates under parametric uncertainty, and maintains an 85\% success rate when subjected to combined wind disturbance and parametric uncertainty. The performance reduction in the most severe scenarios suggests that the UAV is reaching the limits of its control authority. The aggressive turn triggers control saturation because the tail size is kept constant throughout the optimisation, limiting the yaw control moments available to counteract high-amplitude disturbances.

\section{CONCLUSIONS}
\label{conclusion}
This paper presents a robust co-design framework for agile fixed-wing UAVs. Our approach addresses a critical gap in the literature by integrating robustness directly into the joint optimisation process. By embedding stochastic wind disturbances and model uncertainties into the concurrent optimisation of UAV geometry and feedback control laws, our approach establishes a framework for designing agile fixed-wing UAVs that are inherently robust and capable of operating reliably in highly dynamic environments.

Across three agile flight missions, we demonstrate that our robust co-design framework significantly improves system-level performance by enhancing control authority and robustness in stochastic environments. The resulting designs balance task performance and disturbance rejection by tailoring aerodynamic features to specific mission requirements. This includes shifting wing positions to maintain near-neutral stability and high agility, as well as modulating aspect ratio and lateral damping to match the required manoeuvrability. These discoveries are made possible by incorporating nonlinear aerodynamics to accurately capture post-stall behaviour and by utilising quaternion kinematics to provide a singularity-free representation across the entire flight envelope.

While CMA-ES proved effective for navigating these high-dimensional and non-convex design spaces, future iterations could leverage more sample-efficient techniques, such as Bayesian Optimisation \cite{Bjelonic} or Learning to Optimise \cite{Kotary}, to further scale the bi-level architecture. Expanding this framework will involve not just discovering task-specific designs, but exploring complex multi-stage missions and integrating additional aerodynamic, structural, and actuator variables into the co-design loop. We specifically intend to investigate active morphing mechanisms, such as avian-inspired morphing wing platforms \cite{Wuest} and compliant camber morphing structures \cite{Urban}. Ultimately, this work establishes a pathway for designing next-generation agile flight systems that possess the inherent robustness required to navigate unstructured and complex real-world environments.

\bibliographystyle{IEEEtran} 
\bibliography{references}

\end{document}